\pdfoutput=1

\documentclass[11pt]{article}

\usepackage{acl}

\usepackage{times}
\usepackage{latexsym}

\usepackage[T1]{fontenc}

\usepackage[utf8]{inputenc}

\usepackage{microtype}

\usepackage{inconsolata}

\usepackage{algorithm} 
\usepackage{algpseudocode} 

\title{Enhancing Argument Summarization:\\ Prioritizing Exhaustiveness in Key Point Generation and\\ Introducing an Automatic Coverage Evaluation Metric}

\author{
  \textbf{Mohammad Khosravani}\textsuperscript{\rm 1},
  ~\textbf{Chenyang Huang}\textsuperscript{\rm 2},
  ~\textbf{Amine Trabelsi}\textsuperscript{\rm 3} \\
    \textsuperscript{\rm 1}Dept. of Computer Science, Lakehead University  \\
    \textsuperscript{\rm 2}Amii, Dept. of Computing Science, University of Alberta \\
    \textsuperscript{\rm 3}Dept. of Computer Science, Université de Sherbrooke \\
    \normalsize{\tt{mkhosrav@lakeheadu.ca} \, \tt{chenyangh@ualberta.ca} \, \tt{amine.trabelsi@usherbrooke.ca}} 
}

\usepackage{graphicx}

\begin{document}
\maketitle
\begin{abstract}
The proliferation of social media platforms has given rise to the amount of online debates and arguments. Consequently, the need for automatic summarization methods for such debates is imperative, however this area of summarization is rather understudied. The  Key Point Analysis (KPA) task formulates argument summarization as representing the summary of a large collection of  arguments in the form of concise sentences in bullet-style format, called key points. A sub-task of KPA, called Key Point Generation (KPG), focuses on generating these key points given the arguments. This paper introduces a novel extractive approach for key point generation, that outperforms previous state-of-the-art methods for the task. Our method utilizes an extractive clustering based approach that offers concise, high quality generated key points with higher coverage of reference summaries, and less redundant outputs. In addition, we show that the existing evaluation metrics for summarization such as ROUGE are incapable of differentiating between generated key points of different qualities. To this end, we propose a new evaluation metric for assessing the generated key points by their coverage.  Our code can be accessed online.\footnote{\href{https://github.com/b14ck-sun/arg-sum}{github.com/b14ck-sun/arg-sum}}

\end{abstract}

\section{Introduction}
Controversial issues that divide people are prevalent on social media platforms. These platforms provide fertile ground for individuals to express their arguments and engage in debates, leading to a vast amount of textual opinions expressed in diverse manners. Whether it's policymakers, businesses, or individuals, every decision-maker can benefit from a synthesized version that summarizes the main distinct key points related to an issue. 

A key point (KP) should be a concise, high-level argument that aligns with a significant facet of a recurrent argument, while still being informative. Consider the topic of child vaccination. A valid key point might be, ``Routine child vaccinations, or their side effects, are potentially dangerous.''. Another distinct key point could be, ``The decision of vaccinating a child should rest with parents, not the state.'' In this study, we aim to group a multitude of diversely expressed and sometimes poorly formulated raw arguments based on their facets, and to extract those that correspond to valid key points for each facet. Consequently, the summary, i.e., the collection of extracted arguments, should not contain redundancy and should cover the primary key points relevant to the topic of interest.

The work of \citet{bar-haim-etal-2020-arguments} and \citet{friedman-etal-2021-overview} were the first to formulate the Key Point Analysis (KPA) task and to introduce a corresponding dataset, the ArgKP dataset.
The KPA shared task was comprised of two tracks, Key Point Matching (KPM) and Key Point Generation (KPG). The goal of KPM is to map the correct key point to arguments, i.e. predicting whether argument-key point pairs match or not. In KPG, the goal is to generate the key points, given the arguments as inputs. In the KPA task, the term "generated key points" refers to output summary generated by the model. The model itself, whether abstractive or extractive, is referred as a KPG model \cite{friedman-etal-2021-overview}.

The top two previous approaches to KPG, as evaluated by the task organizers, were \citet{bar-haim-etal-2020-quantitative} and \citet{alshomary-etal-2021-key}. However, due to their construction that prioritizes the most popular recurrent arguments, both approaches often generate text that emphasizes \textit{redundant} key points.
Additionally, both methods rely on filtering low-quality arguments, which limit the input. 
This compromises the potential \textit{coverage} of most or all key points, as less popular ones have a limited distribution, and further filtering may further diminish their chances of representation.
Moreover, \citet{bar-haim-etal-2020-quantitative} select only short arguments for inclusion in a summary.
This often yields only a few arguments from an input of around two hundred and produces no summary for inputs of a hundred or fewer.

In this work, we introduce a novel and simple framework for key point generation. It is versatile, catering to both small and large numbers of input arguments. Most critically, it generates a summary comprised of self-contained arguments that cover the majority of existing key points (in a reference summary) with minimal redundancy. In this work we use key point generation and argument summarization interchangeably as the goal of the KPG model is to summarize a group of arguments.

On another aspect, the KPA task organizers relied on human judges to score and rank the KPG models. Judges were instructed to score output summaries based on their coverage of reference summaries and redundancy. Although human evaluation is accurate, it is neither scalable nor reproducible. Alternatively, automatic evaluation metrics, such as ROUGE \cite{lin-2004-rouge}, are often used for evaluating generated summaries. However, as shown by our experiments, they are not as effective for evaluating KPG task. 
Hence, in addition to the argument summarization approach, we propose a new evaluation metric for coverage based on the KPA shared task.

For the \textbf{summarization of arguments}, we employ an extractive clustering based approach. Using semantic-based clustering, we group similar arguments together and use a matching model to find the argument that best represents the cluster. The representatives of clusters are chosen as the summary. This approach allows less popular key points, i.e. smaller clusters, to be represented. Moreover, the matching model used in argument selection eliminates the need for a filtering step that not only limits the arguments that can be extracted, but also introduces additional error and significantly slows down the process. We also demonstrate the versatility of our model by comparing it to previous approaches on datasets other than the traditional ArgKP, which the earlier methods primarily utilized.

To \textbf{automatically assess the coverage} of argument summaries, we utilize a key point matching (KPM) model to compute a coverage score. This model determines whether a given produced argument matches with a reference or ground truth key point. The coverage score quantifies the proportion of matched reference key points in the summary. Furthermore, by enforcing a limit on the number of generated key points, a summary with high coverage of distinct key points would indicate low redundancy. Our experiments confirm the efficacy and appropriateness of the proposed metric compared to traditionally employed ones.

In short, our contributions are as follows:
\begin{enumerate}
    \item  A novel and simple extractive framework for argument summarization based on clustering, that generates concise high quality summary with more coverage and less redundancy compared to existing state-of-the-art work. 
    \item An evaluation metric for assessing summary coverage of the KPG task, that better correlates with the actual key point coverage of summaries compared to existing metrics.
\end{enumerate}

\section{Related Works}
\subsection{Argument Summarization Prior to the KPA Task}
Prior to the introduction of the KPA task, the field of argument summarization was rather underdeveloped, both in terms of available datasets and techniques. However, several research focused on related experiments, such as clustering or extraction of arguments. 

\citet{misra-etal-2016-measuring} aimed to extract different aspects of arguments, called argument facets, similar to key points. The proposed framework first extracted the sentences that include arguments, and further ranked extracted sentences by their similarity to each other, with similar arguments representing an argument facet.
\citet{Trabelsi_Zaïane_2019} developed an approach for unsupervised extraction of argument facets as phrases. These were used to derive arguments articulated in full sentences that reflected contrasting viewpoints or stances. They proposed a phrase topic model that leverages reply-interactions in online debates, enabling the effective clustering of viewpoints and the organized presentation of arguments.
\citet{ajjour-etal-2019-modeling} focused on frame identification using clustering, where a frame refers to arguments that cover the same aspect. The method first clusters the arguments into topics then removes topic-specific features from the arguments, and lastly clusters the “topic-free" arguments into frames. 
\citet{reimers-etal-2019-classification} similarly focuses on argument clustering and classification, however unlike the previous approach they use BERT \cite{devlin-etal-2019-bert} and ELMo \cite{peters-etal-2018-deep} instead of TF-IDF and LSA \cite{RePEc:bla:jamest:v:41:y:1990:i:6:p:391-407}. The authors experimented with different classification and clustering methods in supervised and unsupervised settings.
\subsection{Argument Summarization Following the Introduction of the KPA Task}
The KPA task was based on \citet{bar-haim-etal-2020-arguments}'s work which introduced a dataset of arguments and their reference summaries in the form of key points. The authors showed that a concise list of key points can cover the majority of the arguments covered.  \citet{bar-haim-etal-2020-quantitative} and \citet{alshomary-etal-2021-key} were the top performing models in the KPA task for both key point matching and key point generation tracks. 

Our approach is inspired by \citet{bar-haim-etal-2020-quantitative} where the authors propose a ranking-based extractive method for key point generation. Specifically, first quality candidates are extracted from the input arguments. Next the candidates are ranked by the number of arguments they match to using a key point matching model. Lastly, candidate arguments are selected as summary if their similarity score with previously selected arguments is below a certain threshold. Similarly, we also select the highest coverage arguments as summaries, however, our method differs in the candidate selection, as we don’t rely on selecting high-quality arguments as candidates.

The other top-performing model in the KPA task was proposed by \citet{alshomary-etal-2021-key}, which proposed an extractive method using PageRank, where the nodes are arguments and edges are the matching score between argument pairs. The mentioned approach removes low-quality arguments in the input and only establishes an edge if the matching score is above a threshold. Top nodes according to their importance score are selected as the reference summary, given their similarity score with previously selected nodes is below a threshold.


There has also been a number of research on key point generation following the introduction of the task.
\citet{li-etal-2023-hear} propose a two-step abstractive summarization method based on clustering that utilizes large language models. However, unlike our approach that extracts the best argument in each cluster, they input all arguments belonging to a cluster into a language model for summarization. The proposed method uses BERTopic \cite{grootendorst2022bertopic} for clustering, specifically it uses a pre-trained language model \cite{reimers-gurevych-2019-sentence} for getting contextualized embeddings, performs dimensionality reduction using UMAP \cite{McInnes2018}, and applies HDBSCAN \cite{McInnes2017} to cluster embeddings. In the second step, it further fine-tunes an 11B parameter pre-trained language model \cite{chung2022scaling} to generate the final summary, which is resource-intensive during both training and inference phases.

\section{Methodology}
\subsection{Key Points Generation Models} \label{method_overview}
\begin{figure}[htp]
    \centering
    \includegraphics[width=7.75 cm]{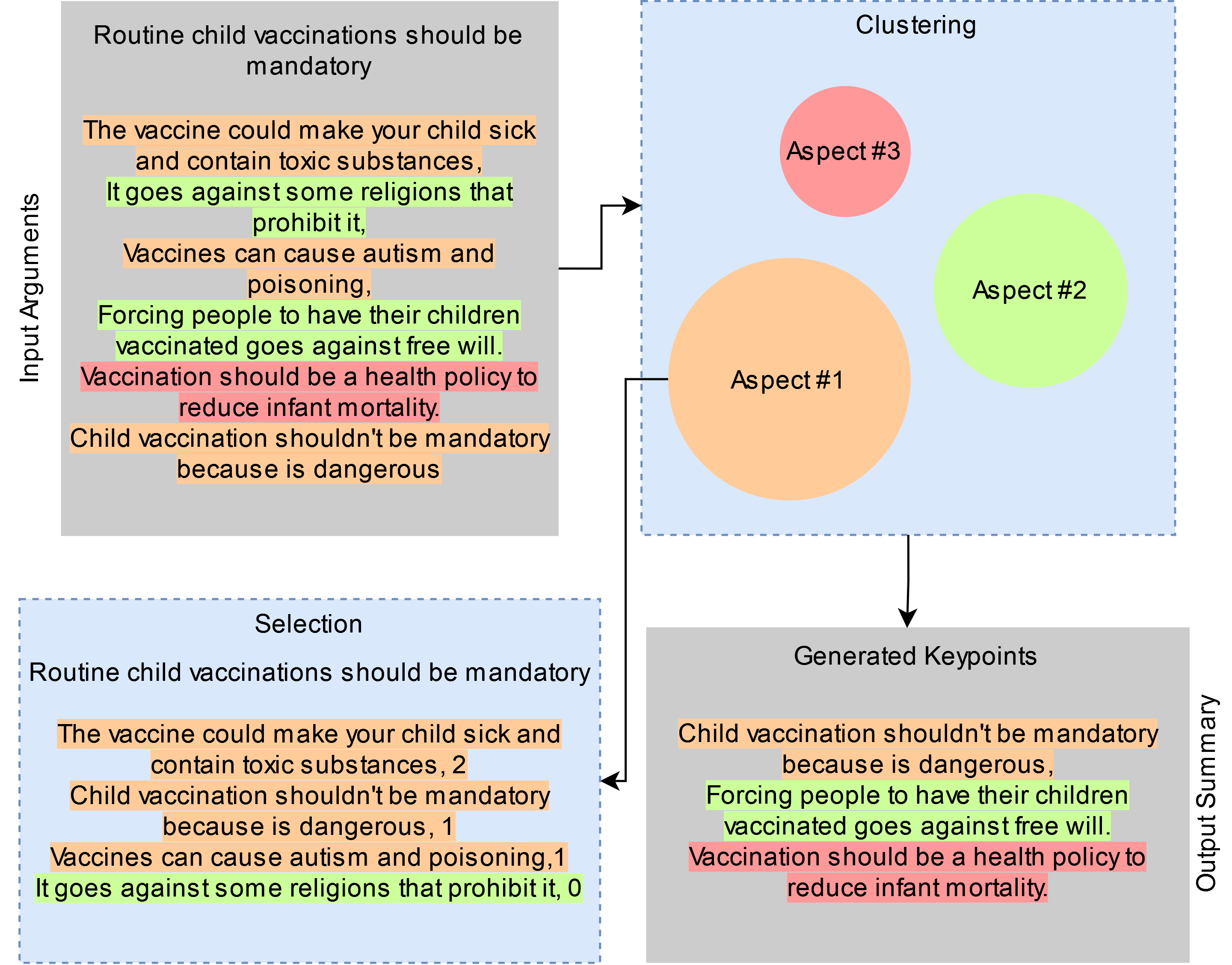}
    \caption{Visual depiction of the proposed KPG method. The color of arguments represents their key point or aspect. First the clustering step groups similar arguments. Next, the selection step chooses the argument with the highest score within each cluster as the cluster representative. The final summary aggregates the representatives from each cluster}
    \label{fig:f1}
\end{figure}

This section describes the proposed framework and our approach to KPG. The input corpus consists of a list of arguments related to a specific topic, such as ``Routine child vaccinations should be mandatory.'' Unlike the original task, the input data doesn't provide any stance information on an argument w.r.t. the topic (e.g., support or oppose). Having this stance information would aid in initially grouping arguments based on their stance, facilitating the separation of lexically similar key points with contrasting viewpoints. However, stance information is often absent in real-world social media texts, and our goal is to devise an effective method for this scenario.
In order to generate summaries, our model first clusters the input arguments, and chooses the best representative sentence in each cluster as a candidate. The  collection of candidates representing clusters is referred to as the generated key points (i.e. output summary). Figure \ref{fig:f1} shows an overview of the approach.

\subsubsection{Clustering}
Previous approaches to KPG such as \citet{bar-haim-etal-2020-quantitative} and \citet{alshomary-etal-2021-key} generate summaries from the input arguments, by favoring popular aspects. This leads to generated summaries that often over-represent some key points, while not mentioning others. This phenomenon occurs because there is often an imbalance in the number of arguments discussing different aspects or key points of a topic. For example, on the topic of “Routine child vaccinations should be mandatory”, there is a total of nine reference or ground truth key points, in the ArgKP dataset \cite{bar-haim-etal-2020-arguments}, where the most popular key point, has fifty arguments discussing it, whereas the least popular key point has only four. 
Our approach aims to represent all the key points equally, prioritizing but not favoring popular ones. This generates a summary with a higher coverage of reference key points and a lower redundancy between generated arguments.

To achieve this goal, we cluster arguments, i.e. sentences, discussing similar key points together. To represent the arguments, we use Sentence-BERT \cite{reimers-2019-sentence-bert} to generate embeddings. Then, we employ Agglomerative Clustering (AC) \cite{aggclust} to group them.
We found that even the best performing pre-trained Sentence-BERT model (all-mpnet-base-v2) used for embedding leads to low Rand index scores, which means the clustered embeddings do not represent the actual clusters. To improve this, we fine-tuned the Sentence-BERT model on the ArgKP dataset using both contrastive loss following \citet{alshomary-etal-2021-key} and Cosine Similarity loss. We used argument-key point pairs in ArgKP dataset as sentence pairs, and their matching label as labels for fine-tuning the Sentence-BERT model.

We observed that fine-tuning the language model improves the Rand index score in the clustering step by more than a 100\%, and that the model fine-tuned using Cosine Similarity outperformed the one fine-tuned using Contrastive loss (Appendix \ref{sec:appendix4}). 
We selected Agglomerative Clustering based on its superior Rand index scores. Additionally, \citet{reimers-etal-2019-classification} identified it as the optimal clustering algorithm for argument clustering and classification. Notably, Agglomerative Clustering doesn't require specifying the number of clusters as a hyperparameter, which is advantageous since this information is frequently unknown during inference. Nevertheless, Agglomerative Clustering remains adaptable to cases where the number of clusters can be estimated.

\subsubsection{Selection} \label{Selection}
After clustering the arguments, we select one argument from each cluster that best represents the entire cluster (see Algorithm \ref{alg:1}). This step is necessary to ensure that the selected arguments cover as many key points as possible. Ideally, the clustering step should cluster all arguments regarding a key point together. In which case, selecting an arbitrary argument from each cluster would be a true representation of that cluster. However, the clustering step does not reflect the actual or correct clusters perfectly. Hence, we need to select the most representative argument from each cluster. 
To achieve this, we sort the clusters by size to prioritize the more popular topics. This approach facilitates the extraction of the most prominent arguments first, from the top clusters, especially when the number of output arguments is constrained or limited to a specific count $k$.  

To select the most representative argument from each cluster, a \textbf{Key Point Matching (KPM)} model, a fine-tuned BERT model in our case, is employed to predict matches between all possible argument-pairs within each cluster. The argument with the highest number of matches in each cluster is selected as the representative (See Algorithm \ref{alg:1} and Algorithm \ref{alg:2}). In the case of a tie in the number of matches (i.e. when the matching model finds more than one argument that best represents the cluster) we select the shortest argument to prioritize conciseness. 
The mentioned BERT model is fine-tuned for entailment on the ArgKP dataset, where argument-key point pairs are given as the input, and the matching label is used as the class (0 for non-matching and 1 for matching). During inference, argument-key point pair are considered matching if the predicted label is 1 (Appendix \ref{sec:appendix_imp_det}). A match between two arguments signifies that one argument entails the other. Consequently, the argument with the highest number of matches in each cluster, i.e. the representative, entails the most number of arguments inside its cluster. 
We refer to this approach as \textbf{Selection with Matching Model} (\textbf{SMM}).

While the generated summaries using the above selection method offer a high coverage of reference summaries, they often included long arguments, lengthening the overall summary. To this end, we, propose an alternative selection method, \textbf{Selection with Scoring Function} (\textbf{SSF}), where a specific scoring function is used.  It takes both the coverage and overall length of an argument into account.
The scoring function gives a score to an argument $arg$ in each cluster $c$ according to the number of matches while penalizing lengthy sentences as presented in line 5 of Algorithm \ref{alg:2}.
%
It uses
$i$ 
an exponent hyperparameter 
which
determines the importance of matches.
It controls the scoring function's emphasis on shorter sentences when candidates have similar match numbers.

\begin{algorithm}
	\caption{Argument Selection} 
    \hspace*{\algorithmicindent} \textbf{Input:} Clusters $C$ (sorted largest to smallest),\\
    \hspace*{\algorithmicindent} Selection Function (SMM or SSF): $SF$,\\
    \hspace*{\algorithmicindent} \textbf{Output:} Generated Key Point $GKP$ 
	\begin{algorithmic}[1]
		\For {$c\ in\ C$}
			\For {$arg\ in\ c$}
                \State $c^\prime\ \gets c.remove(arg)$
				\State $Score_{c, arg} \gets SF(c^\prime, arg)$
			\EndFor
			\State $Selected_c \gets argmax_{arg} (Score_{c, arg})$
		\EndFor
        \State {$GKP \gets [Selected_{c1}, \ldots,Selected_{|C|}]$}
        \State \Return {$GKP$}
	\end{algorithmic} 
 \label{alg:1}
\end{algorithm}

\begin{algorithm}
	\caption{Selection Function: SMM \& SSF} 
    \hspace*{\algorithmicindent} \textbf{Input:} Cluster $c$, Argument $arg$,\\ \hspace*{\algorithmicindent} Exponent Hyperparameter $i$,\\
    \hspace*{\algorithmicindent} Key Point Matching Model: $KPM$\\
    \hspace*{\algorithmicindent} \textbf{Output:} $Score_{c, arg}$ 
	\begin{algorithmic}[1]
    \Procedure{SMM}{$c, arg$}
    \State \Return {$Score_{c, arg} \gets \textnormal{MATCHES}(c, arg)$}
    \EndProcedure
    \item[]
    \Procedure{SSF}{$c, arg$}
    \State \Return {$ Score_{c, arg} \gets  \frac{{\textnormal{MATCHES}(c, arg)}^i}{\# of words(arg)}$}
    \EndProcedure
    \item[]
    \Procedure{Matches}{$c, arg$}
        \State $matches \gets 0$
        \For {$arg^\prime\ in\ c$}
            \If {$KPM(arg,arg^\prime)$}
                \State $matches++$
            \EndIf
		\EndFor
    \State \Return {$matches$}
    \EndProcedure
    \item[]
	\end{algorithmic} 
 \label{alg:2}
\end{algorithm}

\subsection{Coverage Evaluation Measure}
The standard evaluation measure for summarization tasks, ROUGE \cite{lin-2004-rouge}, is not effective for evaluating the KPG task. In addition to its reported problems on reproducibility and comparability \cite{grusky-2023-rogue}, ROUGE score is concerned with the n-grams overlap between candidates and summaries, and fails to consider the semantics of documents \cite{li-etal-2023-hear}. This makes the ROUGE score less effective for evaluating arguments where the same viewpoint can be expressed with completely different wording and opposing arguments can have similar words.
In the KPG task, an ideal summary should be concise, non-redundant, and cover different aspects or facets of the topic, according to the task organizers. To this end, task organizers used human judges to evaluate the generated summaries based on how redundant a summary is, and how it captures central points to the topic. Inspired by this, we propose an automatic evaluation metric for the KPG task, a coverage measure. However, unlike the KPA task where the goal was to cover the most important key points, we argue that covering more aspects of a topic is more important as it leads to a more comprehensive summary. It is also equally important to have a summary with a low number of duplicates, i.e. arguments that cover the same aspect or key point. 
The coverage measure assesses the extent to which different aspects of the topic are represented.

When the number of generated key points (i.e. output arguments) in a summary is capped or limited to a ground truth number of arguments, a  high key point coverage value signifies that most arguments are distinct, thereby implying minimal redundancy.

\subsubsection{Coverage Computation} \label{Coverage_Computation}
The coverage measure predicts the percentage of actual ground truth key points captured by the produced summary. Our proposed coverage metric computes a coverage score by pairing each generated key point with every reference key point and use a matching model to predict if the argument-key point pair is a match. The coverage score is calculated as the percentage of all reference key points covered. As for the matching model, any classification method, such as the ones submitted for the key point matching track of the KPA task can be used, since the goal is to predict matching argument-key point pairs. Additionally, the BLEURT and BARTScore models used by \citet{li-etal-2023-hear} also evaluate coverage since the proposed soft-precision metric matches each reference key point with the highest scoring generated key point (and vice-versa for soft-recall).
Our proposed key point matching uses the same fine-tuned BERT model in Section \ref{Selection}, similar to \citet{bar-haim-etal-2020-quantitative}. We frame the KPM problem as an entailment problem unlike some previous  approaches that use cosine similarity between argument and key point embeddings. Details on implementation settings can be found in Appendix \ref{sec:appendix_imp_det}.


\begin{table*}[t]
    \centering
    \begin{tabular}{cccccc}
    \hline
        ArgKP & Coverage & Redundancy & R1 & R2 & RL \\ \hline
        Method (SMM) & \textbf{59.59\%} & 2.46\% & 0.152 & 0.026 & 0.132 \\ 
        Method (SSF) & 57.67\% & \textbf{2.27\%} & 0.158 & 0.026 & 0.141 \\ 
        Alshomary w SS & 49.09\% & 3.41\% & 0.195 & \textbf{0.032} & 0.175 \\ 
        Alshomary & 45.45\% & 3.23\% & \textbf{0.202} & 0.028 & \textbf{0.186} \\ 
        BarHaim & 37.67\% & 5.26\% & 0.153 & 0.028 & 0.136 \\ \hline
    \end{tabular}
\end{table*}

\begin{table*}[t]
    \centering
    \begin{tabular}{cccccc}
    \hline
        Debate & Coverage & Redundancy & R1 & R2 & RL \\ \hline
        Method (SMM) & \textbf{72.38\%} & \textbf{1.42\%} & 0.06 & 0.005 & 0.052 \\ 
        Method (SSF) & 57.98\% & 1.86\% & 0.079 & 0.007 & 0.069 \\ 
        Alshomary & 64.88\% & 1.68\% & 0.068 & 0.003 & 0.062 \\ 
        BarHaim & 56.41\% & 1.74\% & \textbf{0.084} & \textbf{0.015} & \textbf{0.08} \\ \hline
    \end{tabular}
    \caption{Actual coverage, redundancy, and ROUGE scores for each model's output. The coverage and redundancy are computed using the labeled data. SS refers to separate stances. Numbers in bold represent the best results.}
\label{table:1}
\end{table*}
\subsubsection{Coverage Datasets} \label{coverage_datasets}
To test the performance of different evaluation metrics, we created a set of pseudo summaries from the ArgKP test set, with different levels of coverage, 100\%, 75\%, and 50\%. We named them Coverage Datasets -- a dataset for each level of coverage. These pseudo summaries each contain the same number of arguments i.e. 25, sampled from the unseen ArgKP test set.
Each coverage dataset, only covers its respective percentage of key points from the ArgKP test set. For example, for the 50\% coverage dataset, we randomly select 50\% of the key points on each topic (e.g., child vaccination), and only include arguments from those key points, with each selected key point having at least one argument related to them. To ensure fairness, the total number of arguments in each dataset is equal. Additionally, each coverage dataset was randomly sampled ten times. As an example the 100\% coverage dataset, is randomly sampled 10 times, each time with different 25 arguments that cover 100\% of the key points. Any results reported for coverage datasets are averaged over all ten instances.
We utilize these coverage datasets to evaluate the efficacy of various evaluation metrics by tasking them to estimate the coverage provided by these datasets. 

\section{Experimental Setup}
\subsection{ Experiments on the KPG Method}
\textbf{Overview:} We perform both automatic and human evaluations to prove the effectiveness of our method. 
To showcase the ability of the proposed method to generate superior summaries, we compare the generated key points to previous approaches. Mainly we compare our work to the top two performing models in KPA tasks, \citet{alshomary-etal-2021-key} and \citet{bar-haim-etal-2020-quantitative}\footnote{\href{https://early-access-program.debater.res.ibm.com}{early-access-program.debater.res.ibm.com}}. We experiment on both versions of our selection method discussed in Section \ref{Selection}: SMM and SSF. We evaluate and compare different methods based on their coverage, redundancy, conciseness, and argument quality.
For comparison, we limit the number of generated outputs by the models to the number of key points in the reference (ground truth) datasets.
This was done as different methods produce summaries with different numbers of key points, and longer summaries are more likely to have a higher coverage of reference key points.
Additionally, we have grouped all arguments that do not belong to any existing key point in the ArgKP dataset under one shared key point. This ensures that arguments that do not belong to any key point also contribute to coverage score. 

\textbf{Coverage and Redundancy:} In order to evaluate the actual coverage, we calculate the percentage of reference key points covered by the generated key points. As all the methods are extractive, the true key point for each generated key point is known, which allows us to calculate the effective and actual percentage of reference key points covered. It is important to note that in this experiment we calculate the actual coverage using labeled data, and not the predicted coverage computed by our proposed metric. For redundancy, we consider the generated key points that belong to the same reference key point as duplicates and count the number of duplicates in each model, and present the percentage of duplicates (number of duplicate pairs divided by the number of all possible pairs).

\textbf{Conciseness and Quality:} In order to judge the quality of generated key points, we asked human judges to score those produced by different methods with respect to their reference key points. While this evaluation might seem unnecessary since the proposed and previous approaches are extractive, these approaches modify the structure of arguments by splitting multi-sentence arguments into single sentences which could affect their clarity. Inspired by human evaluation done by \citet{li-etal-2023-hear}, we ask the judges to score argument-key point pairs (+1 for matches, -1 non-matches, 0 for not sure) based on two conditions: 1. The argument covers the same aspect as the key point. 2. The argument is a clear and understandable argument regarding an aspect by itself given the topic (complete instructions are provided in Appendix \ref{sec:appendix_hum_ins}). We paired each  generated key point with its respective matching key point based on the ArgKP dataset. For example, on the topic "Is US a good country to live in?" the generated key point "it doesn't provide it's citizens with free healthcare", was paired with the reference key point "The US has unfair health and education policies".

We sampled a total of 20 generated key points from four topics, two topics from the Debate dataset (Section \ref{dataset}) and two from ArgKP, for each model. We asked three graduate students to score each pair and averaged the score across judges for each pair. Lastly, we compare the average number of words per generated key point over all topics in the ArgKP test set to showcase the ability of the proposed method to generate concise outputs.

\begin{table*}[t]
    \centering
    \begin{tabular}{ccccccc}
    \hline
        ArgKP & Coverage Measure & BLEURT & BARTScore & R1 & R2 & RL \\ \hline
        100\% & \textbf{83.18\%} {\tiny\textpm 0.052} & 66.26\% {\tiny \textpm 0.006} & 60.29\% {\tiny \textpm 0.044}& 0.166 {\tiny \textpm 0.007}& 0.032{\tiny \textpm 0.005} & 0.150 {\tiny \textpm 0.007}\\ 
        75\% & \textbf{77.27\%} {\tiny \textpm 0.056} & 61.11\% {\tiny \textpm 0.015}& 59.26\% {\tiny \textpm 0.023}& 0.169 {\tiny \textpm 0.010}& 0.033 {\tiny \textpm 0.006}& 0.153 {\tiny \textpm 0.010}\\ 
        50\% & 64.84\% {\tiny \textpm 0.055}& \textbf{56.38\%} {\tiny \textpm 0.017}& 57.61\% {\tiny \textpm 0.029}& 0.165 {\tiny \textpm 0.011}& 0.033 {\tiny \textpm 0.006}& 0.151 {\tiny \textpm 0.011}\\ \hline
    \end{tabular}
\end{table*}

\begin{table*}[t]
    \centering
    \begin{tabular}{ccccccc}
    \hline
        Debate & Coverage Measure & BLEURT & BARTScore & R1 & R2 & RL \\ \hline
        100\% & 82.08\% {\tiny\textpm 0.067}& 91.67\% {\tiny\textpm 0.015}& \textbf{92.71\%} {\tiny\textpm 0.015}& 0.064 {\tiny\textpm 0.002}& 0.005 {\tiny\textpm 0.000}& 0.056 {\tiny\textpm 0.001}\\ 
        75\% & \textbf{76.67\%} {\tiny\textpm 0.040}& 90.63\% {\tiny\textpm 0.015}& 91.67\% {\tiny\textpm 0.015}& 0.066 {\tiny\textpm 0.003}& 0.006 {\tiny\textpm 0.001}& 0.057 {\tiny\textpm 0.002}\\ 
        50\% & \textbf{70.42\%} {\tiny\textpm 0.063}& 89.58\% {\tiny\textpm 0.059}& 94.79\% {\tiny\textpm 0.044}& 0.064 {\tiny\textpm 0.003}& 0.006 {\tiny\textpm 0.001}& 0.056 {\tiny\textpm 0.003}\\  \hline
    \end{tabular}
    \caption{Different metrics for coverage prediction on coverage datasets (100\%, 75\%, 50\%). The numbers in bold represent the best results (closest prediction to actual coverage). Values in subscripts represent standard deviation.}
\label{table:2}
\end{table*}
\subsection{Experiments on Evaluation Metrics}
\textbf{Coverage Prediction:} We assess the performance of our coverage measure compared to other metrics for KPG, namely ROUGE and \citet{li-etal-2023-hear}. We assessed the metrics by tasking the evaluation metrics and models to predict the coverage of different Coverage datasets (Section \ref {coverage_datasets}). This allowed us to assess how good the evaluation metrics are at correlating an effective high coverage and low redundancy summary to a high score. First, we computed the ROUGE score on the datasets, with the key points as reference summaries. 

Second, we use the evaluation metric introduced by \citet{li-etal-2023-hear} for computing Soft-Precision and Soft-Recall. The proposed Soft-Precision aims to finds the reference key point with the highest similarity score for each generated key point and vice-versa for Soft-Recall. Soft-F1 is the harmonic mean between soft-precision and soft-recall. The similarity score is calculated by BLEURT and BARTScore models, given the generated and reference key points as inputs. The Soft-F1 aims to evaluate the validity, sentiment, informativeness, single-aspect, significance, \textbf{coverage}, faithfulness, and redundancy of generated key points.
In our experiment, we assigned every generated key point to the reference key point with the highest matching score, similar to the authors' approach, which evaluates the effectiveness of models at correctly assigning key points to arguments. We compared the predicted coverage of each metric to the actual coverage.

\textbf{On Generated Summaries:} We also evaluated summaries generated by different models using ROUGE, soft-F1 scores \citet{li-etal-2023-hear}, and the proposed coverage measure. However, due to Soft-F1 metric limitation, where the number of generated key points in the output should match the number of reference key points, we were not able to evaluate the outputs generated by every model.

\subsection{Dataset} \label{dataset}
The experiments are done on the dataset proposed for the KPA task, the ArgKP dataset \cite{friedman-etal-2021-overview} with ~24K argument/key-point pairs. The dataset contains 24 topics for train, and 4 for development, with ~800 arguments and 3 topics for test set. 
To further prove the effectiveness and generalizability of our approaches, we compared our method and evaluation metric on the Debate dataset \cite{hasan-ng-2014-taking}. The Debate dataset contains arguments and aspects, similar to the ArgKP dataset, on four topics, with ~3K argument/key-point pairs.
To assess our proposed argument summarization approach, we compared it with \citet{bar-haim-etal-2020-quantitative}, and \citet{alshomary-etal-2021-key} on the Debate dataset.
To assess our coverage evaluation measure, we compared it to BLEURT and BARTScore for coverage prediction. 
To this end, we generated a set of different coverage datasets from the Debate dataset, similar to the previous coverage dataset (Section \ref{coverage_datasets}) with 75 arguments for each dataset, as the Debate dataset has more arguments per topic.

\begin{table*}[!ht]
    \centering
    \begin{tabular}{cccccccc} \hline
        ArgKP & Coverage & Redundancy & F1 BL & F1 BS & R1 & R2 & RL \\ \hline
        Method (SMM) & 35.75\% (\textbf{59.59\%}) & (2.46\%) & 49.01\% & 57.94\% & 0.152 & 0.026 & 0.132 \\ 
        Method (SSF) & \textbf{39.89\%} (57.67\%) & (\textbf{2.27\%}) & 52.84\% & 60.81\% & 0.158 & 0.026 & 0.141 \\ 
        Alshomary & 30.70\% (45.45\%) & (3.23\%) & \textbf{54.58\%} & \textbf{63.04\%} & \textbf{0.202} & \textbf{0.028} & \textbf{0.186} \\ 
        BarHaim & 23.53\% (37.67\%) & (5.26\%) & N/A & N/A & 0.153 & \textbf{0.028} & 0.136 \\  \hline
    \end{tabular}
    \caption{The predicted coverage of models, alongside the Soft-F1 scores using BLEURT (BL) and BARTScore (BS), and ROUGE. The numbers in parenthesis represent the actual coverage and redundancy. Numbers in bold represent the best output.}
\label{table:4}
\end{table*}

\section{Results and Analysis}
\subsection{Evaluation of the KPG Method}
\subsubsection{Coverage and Redundancy}
Our method outperforms both \citet{bar-haim-etal-2020-quantitative} and \citet{alshomary-etal-2021-key} methods in producing more exhaustive and less redundant key points on the ArgKP dataset. In addition, we compare the output of our model, with Alshomary's output on the original KPA task when arguments are separated by their stances (Alshomary w SS). 
However, we cannot apply BarHaim's method similarly, as it requires more than 100 arguments per input, and each set of arguments separated by stance has fewer.
Table \ref{table:1} shows the coverage, redundancy and the ROUGE score of the proposed method, BarHaim, and Alshomary on the ArgKP and the Debate datasets with respect to the reference (true) key points. Our experiments show that both proposed versions of our approach, SMM and SSF, outperform other methods on the ArgKP dataset. The proposed methods are also generalizable to the Debate dataset as the method with SMM also outperforms both other methods and the method with SSF outperforms BarHaim on the Debate dataset. It is also important to note that the inferior performance of BarHaim in ArgKP dataset is partly due to the fact that it has a maximum length limit on the extracted key points, which makes the model generate less key points. However, this limit is fine-tuned for this dataset and cannot be changed or modified. 
On another aspect, ROUGE scores show no correlation with coverage and redundancy of outputs. We examine ROUGE's performance in more detail in Sections \ref{eval_covs} and \ref{eval_covs_gkp}.

\subsubsection{Conciseness and Quality}
To showcase the proposed method's ability to generate quality short sentences, we compare the average word length and the human-evaluated quality scores of generated key points. Table \ref{table:3} shows the scores averaged over all outputs. 
The results show our method surpasses Alshomary's in conciseness, yet BarHaim's approach, which only extracts short arguments, still yields briefer key points.

\begin{table}[!ht]
    \centering
    \begin{tabular}{ccc} \hline
        Method & Avg. Words & Arg. Quality \\ \hline
        Method (SMM) & 18.6 & \textbf{0.63} \\ 
        Method (SSF) & 12.4 & 0.4 \\ 
        Alshomary & 15.3 & 0.46 \\ 
        BarHaim & \textbf{6.8} & 0.23 \\  \hline
    \end{tabular}
    \caption{Average number of words and quality scores per generated key point, averaged for each model.}
    \label{table:3}
\end{table}

Moreover, the human judges found the proposed methods' outputs more understandable, as the best with method with SMM having the highest score and method with SSF having a similar score to the Alshomary's method. The human scores also indicate that longer outputs are easier to understand. The Krippendorff’s $\alpha$ for inter-annotator agreement is 0.47 across all topics, with a score of 0.53 on the ArgKP dataset, and 0.40 on the Debate dataset. 

\subsection{Evaluation of the Coverage Measure} \

\subsubsection{Comparison of the Evaluation Metrics} \label{eval_covs}
We compared the performance of our coverage measure to ROUGE and the method proposed by \citet{li-etal-2023-hear}. For comparison, we used the proposed coverage measure, BLEURT, and BARTScore, to predict the coverage of different coverage datasets. Additionally, we calculated the ROUGE score of coverage datasets to examine ROUGE's ability to score outputs of different quality but the same length. Table \ref{table:2} shows the predicted coverage of datasets using all evaluation metrics.

Our experiments show that the ROUGE score is not capable of differentiating between datasets with different levels of coverage. The BARTScore performs only slightly better, as the difference in predicted coverages is insignificant. BLEURT, however, performs reasonably well on the ArgKP dataset, but it does not generalize as well to the Debate dataset. We hypothesize the ineffectiveness of BLEURT and BARTScore is due to the fact that they are not trained on argumentative text, which uses a different distribution of words. ROUGE's poor performance, on the other hand, is the result method's inability to take the semantics of sentences into account, this is especially true in the argument domain as arguments on the same topic belonging to different key points often use similar words, which is not handled effectively by ROUGE. The inaccurate coverage prediction of BLEURT and BARTScore indicates that the models are not capable of assigning the generated key points to the correct reference key points, making them ineffective at predicting soft-F1 score. In contrast, the proposed approach's predictions are closer to the actual coverage of datasets. 
Also, 
the metric's effectiveness extends to unseen datasets, as evidenced by its performance on the Debate dataset.

Lastly, our KPM model used in the coverage metric (see  Section \ref{Coverage_Computation}) demonstrates high accuracy on the ArgKP dataset.
On the ArgKP test set, it scores a 90.01\% micro percision, compared to the 87.79\% of the best performing KPM model in the KPA task by \citet{alshomary-etal-2021-key}.

\subsubsection{Evaluation of the Coverage Metrics on the Generated Outputs} \label{eval_covs_gkp}
We further assessed the generated outputs using ROUGE, our designated metric, and the metrics introduced by \citet{li-etal-2023-hear}. 
Table \ref{table:4} displays the scores predicted by each evaluation metric, alongside the actual coverage and redundancy values. The Soft-F1 score can only be calculated when the number of generated key points is equal to the number of reference key points, making it incompatible with BarHaim. Table \ref{table:4}'s results suggest that our proposed coverage measure often estimates the relative ranking of outputs accurately. On the other hand, alternative evaluation metrics don't consistently link a high coverage/low redundancy output with a top score. Notably, both F1 measures identify Alshomary's output as the best among the evaluated outputs. However, according to Table \ref{table:1} Alshomary's output has a lower coverage of reference key points and higher redundancy in generated key points. Also according to Table \ref{table:3}, it has lower argument quality and comparable length to method with SMM. In this case it is not clear how/why a certain summary is better than the others. 
Therefore, we argue that employing a single metric to compare generated key points overall is not effective or descriptive. This is because generated KPs vary in aspects like coverage, redundancy, length, and quality, each demanding its own evaluation.

\section{Conclusion}
This paper proposes a new extractive method for KPG that aims to generate key points with a high coverage of reference key points and less redundancy. Our method adopts a clustering-based approach with two distinct selection methods to achieve this goal. Our experiments show that the proposed methods outperform previous SOTA in terms of coverage, redundancy, and quality 
. 
Lastly, we proposed a metric for coverage, that outperforms previous metrics on coverage prediction. 

\section*{Limitations}
\textbf{Dataset: }The test set for ArgKP dataset, while relatively small, is the only dataset for the task. However we aimed to verify our  experiments by utilizing another dataset. We also did not filter offensive arguments in the corpus, as a result the generated result may include text which some readers might find offensive. It is important to highlight that detecting inappropriate language wasn't the goal of this study.\\
\\
\textbf{Method \& Evaluation Metrics: } The evaluation metrics, while more accurate than previous approaches, are still not accurate enough to rank the models with 100\% accuracy, when the actual coverage and redundancy of two generated summaries is insignificant. Additionally, the coverage evaluation metric does not factor in the hierarchical entailment relation between arguments themselves or arguments and reference key points. As an example, the model might predict that a general reference key point (e.g. vaccinations may have unwanted side effects) is covered by a specific generated key point (e.g. vaccinations may have a specific side effect). Prioritizing these general key points over specific ones could be a potential future work, which might also require a dataset where the hierarchical relations between arguments are traced or annotated (i.e. which key points cover more fine-grained ones).\\
Additionally, the proposed approach has no way of enforcing generation of relatively equal key points for both stances. This could potentially lead to summaries that favor one stance more often than the other. 

\section*{Ethics Statement}
The dataset used for training/fine-tuning models is an existing anonymized dataset without data protection issues. Moreover, the human annotators for the task were informed that their work would be used, and were unpaid volunteers unrelated to the research. Only trained annotators who understood the task were selected. The only collected data was ratings produced by the annotators.


\section*{Acknowledgements}
This research was partially supported by the Vector Institute through the Vector Scholarship in Artificial Intelligence, by the Natural Sciences and Engineering Research Council of Canada (NSERC) under Grant No. RGPIN-2022-04789, and by Compute Ontario (\href{https://www.computeontario.ca/}{https://www.computeontario.ca/}) and the Digital Research Alliance of Canada (\href{https://alliancecan.ca/}{https://alliancecan.ca/}). We would like to thank the annotators and the reviewers for their valuable feedback.

\bibliography{anthology,custom}

\appendix

\section{Output Examples on Child Vaccination Topic}
\label{sec:appendix1}

Outputs generated by BarHaim
\begin{itemize}
  \item vaccinating children helps eradicate disease
  \item Child vaccination shouldn't be mandatory because is dangerous
  \item to keep schools safe children must be vaccinated
  \item protecting infants must be a priority for all
  \item this vaccine could cause unwanted side effects
  \item vaccination at birth is a human duty
\end{itemize}

Outputs generated by Alshomary
\begin{itemize}
    \item People all around the world vaccinate their children to protect them from any life threatening disease.
    \item Parents should be allowed to choose if their child is vaccinated or not.
    \item Parents don't always know best and failure to vaccinate can be catastrophic for a child.
    \item child vaccinations should be mandatory to provide decent health care to all.
    \item Vaccines in children should not be mandatory because they can have consequences for their health in the future
    \item When vaccines are mandatory, they can infringe on family religious choices
    \item each child must be vaccinated so that they can live more peacefully
    \item childhood vaccination is necessary as it helps the growth of children
    \item child vaccinations is not mandatory because it may cost the country unnecessary funds.
\end{itemize}

Output generated by method with SMM
\begin{itemize}
    \item vaccines must be compulsory for children because in this way we prevent later diseases in infants,
    \item Child vaccination should be mandatory because they may transfer the virus,
    \item someone from the child population could suffer side effects,
    \item No-one can tell a parent that they must vaccinate their child, it is against their human rights,
    \item Child vaccination shouldn't be mandatory because children don't catch the virus,
    \item Children are the future of the country and the world, taking care of them should be our greatest concern so that they grow up as healthy adults that is why vaccination should be mandatory,
    \item to keep schools safe children must be vaccinated,
    \item Vaccines save up to 3 million lives a year, protecting children from life-threatening and highly infectious diseases,
    \item Routine childhood vaccinations should not be mandatory, because the contraindications or what negative effect it may cause the child are not known, the parents should decide,
\end{itemize}

Output generated by method with SFM
\begin{itemize}
    \item Prevents a large number of diseases,
    \item this vaccine could cause unwanted side effects,
    \item Parents should decide what is best for their child.,
    \item Child vaccination should be mandatory to avoid the virus,
    \item Child vaccination shouldn't be mandatory because children don't catch the virus,
    \item protecting infants must be a priority for all,
    \item to keep schools safe children must be vaccinated,
    \item because they can have very dangerous reactions to vaccines,
    \item the vaccine provide inmunity to  the people  and prevents to contract the dissease
\end{itemize}


\section{Rand Index Scores}
\label{sec:appendix4}

Our experiments show that fine-tuning the SBERT model significantly improves the accuracy of clustering step, which improves the overall quality of generated key points. We use Rand index score to measure the accuracy of predicted clusters w.r.t to the original clusters. We also remove all arguments that do not belong to any clusters for this evaluation. Table below shows the Rand index score improvements after fine-tuning the model using both contrastive and cosine similarity loss.
\begin{table}[htp]
    \centering
    \begin{tabular}{cccc}
    \hline
        Rand Index Score & Topic1 & Topic2 & Topic3 \\ \hline
        Before FT & 0.18 & 0.19 & 0.15 \\ 
        FT w CS & 0.43 & 0.45 & 0.39 \\ 
        FT w Cont. & 0.38 & 0.44 & 0.32 \\ \hline
    \end{tabular}
    \label{table:randindexscore}
    \caption{Rand index scores for the clustering before and after fine-tuning on the ArgKP train set using Cosine Similarity loss and Contrastive loss. Topic 1, 2, and 3 refer to topics in ArgKP test set.}
\end{table}

\section{Implementation Details}
\label{sec:appendix_imp_det}
\subsection{Key Point Generation Models}
For embedding generation, we use “all-mpnet-base-v2”, and fine-tune it using Cosine similarity loss the ArgKP dataset. Agglomerative clustering with a distance threshold of 1.5 is used for clustering embeddings. For the argument selector, we used a fine-tuned BERT on the ArgKP dataset. We format each input as “[CLS] argument [SEP] key point [SEP]” for input, and [logit 1, logit 2], where [1, 0] represents class zero or non-matching and [0, 1] represents matching argument-key point pairs, for the output. A pair is considered matching if the second logit is greater than the first one. For the Selection with Scoring  (SSF), we tried different formulas (e.g. the number of matches, MATCHES,  divided by number of words or \(e^{\textnormal{MATCHES}}\) divided by number of words). Empirically, we find that setting the exponent hyperparameter $i$ to 5 (see Algorithm \ref{alg:2}) produces the best result.

\subsection{Evaluation Metric}
We used the same BERT model used in argument selection for coverage, however we make a few modifications. First, we limited the number of reference key points that can be assigned to a generated key point to one. In case of multiple reference key points matching to a single generated key point, we assign the highest scoring reference key point to the generated one. Our experiments show that setting this limit of maximum one improves the accuracy. This is in line with the nature of input arguments as only 4\% of the arguments cover more than one key point. We also experimented with setting different values for limits and thresholds on the coverage model similar to \citet{bar-haim-etal-2020-quantitative}. As for the ROUGE metric, we used Python's rouge\_score library version 0.1.2.


\section{Task Description for Human Evaluators}
\label{sec:appendix_hum_ins}
\textbf{Task}
You are given a pair of argument and key point sentences, and your have to decide whether the pair match or not
A matching pair should satisfy two conditions: 1. The argument should cover the same aspect as the key point. 2. The argument should be a clear and understandable argument regarding an aspect by itself given the topic.
You are supposed to give a score to each argument key point pair
Score 1 if the argument and key point are matching
Score -1 if the argument and key point are not matching
Score 0 if you are not sure 
\textbf{Data}
\begin{itemize}
    \item You are given 80 sentences in total, 20 sentences for 4 topics. The topics are:
    \begin{itemize}
        \item The USA is a good country to live in
        \item Social media platforms should be regulated by the government
        \item Abortion
        \item Gay rights
    \end{itemize}
    \item The arguments and key points can be “pro” the topic, or “against” it
\end{itemize}

\textbf{Examples}
\begin{itemize}
    \item Below are some examples of matching and non-matching pairs
    \item Matching
    \begin{itemize}
        \item Argument: Prevents a large number of diseases
        \item Key point: Vaccines prevent diseases
        \item Matching because “Prevents a large number of diseases” is understable given the topic of “Vaccination should be mandatory”
    \end{itemize}
    \item Non Matching
    \begin{itemize}
        \item Argument: Vaccination should not be mandatory
        \item Key Point: Vaccines have side-effects
        \item Non matching because the key point mentions the aspect of side effect and the argument does not
    \end{itemize}
    \item Non Matching
    \begin{itemize}
        \item Argument: That is not possible.
        \item Key point: Vaccinating everyone is not possible
        \item Non matching because the argument is not an understandable sentence by itself, i.e. it is too vague.
    \end{itemize}
\end{itemize}


\section{Computational Cost}
\label{sec:appendix_computational}
All the experiments on method were done on V100-32gb. Total training and evaluation for the method took about 12 hours, with each run about 30 minutes.

\end{document}